\pdfoutput=1
\documentclass[11pt]{article}

\usepackage[]{acl}

\usepackage{times}
\usepackage{latexsym}

\usepackage[T1]{fontenc}
\usepackage[utf8]{inputenc}

\usepackage{microtype}

\usepackage{amsmath}
\usepackage{caption}
\usepackage{color}
\usepackage{enumerate}
\usepackage{graphicx}
\usepackage{multirow}
\usepackage{subcaption}

\usepackage{CJKutf8}

\title{Building a Dialogue Corpus Annotated with\\Expressed and Experienced Emotions}

\author{Tatsuya Ide \and Daisuke Kawahara \\
  Department of Computer Science and Communications Engineering, Waseda University \\
  \texttt{\{t-ide@toki., dkw@\}waseda.jp} \\}

\begin{document}
\maketitle
\begin{abstract}
In communication, a human would recognize the emotion of an interlocutor and respond with an appropriate emotion, such as empathy and comfort.
% Developing a dialogue system with such a human-like ability is essential.
% In this work, we propose a dialogue corpus with annotation of expressed and experienced emotions toward such a system.
Toward developing a dialogue system with such a human-like ability, we propose a method to build a dialogue corpus annotated with two kinds of emotions.
We collect dialogues from Twitter and annotate each utterance with the emotion that a speaker put into the utterance (expressed emotion) and the emotion that a listener felt after listening to the utterance (experienced emotion).
We built a dialogue corpus in Japanese using this method, and its statistical analysis revealed the differences between expressed and experienced emotions.
We conducted experiments on recognition of the two kinds of emotions.
The experimental results indicated the difficulty in recognizing experienced emotions and the effectiveness of multi-task learning of the two kinds of emotions.
We hope that the constructed corpus will facilitate the study on emotion recognition in a dialogue and emotion-aware dialogue response generation.
\end{abstract}

\section{Introduction}
\label{sec:intro}

Text-based communication has become indispensable as society accelerates online.
In natural language processing, communication between humans and machines has attracted attention, and the development of dialogue systems has been a hot topic.
Through the invention of Transformer \citep{NIPS2017_3f5ee243} and the success of transfer learning (e.g., \citet{radford2018improving, devlin-etal-2019-bert}), the performance of natural language understanding models and dialogue systems continues to improve.
In recent years, there have been studies toward building open-domain neural chatbots that can generate a human-like response \citep{zhou-etal-2020-design, adiwardana2020humanlike, roller-etal-2021-recipes}.

One of the keys to building more human-like chatbots is to generate a response that takes into account the emotion of the interlocutor.
A human would recognize the emotion of the interlocutor and respond with an appropriate emotion, such as empathy and comfort, or give a response that promotes positive emotion of the interlocutor.
Accordingly, developing a chatbot with such a human-like ability \citep{rashkin-etal-2019-towards, Lubis_Sakti_Yoshino_Nakamura_2018, 8649596} is essential.
Although several dialogue corpora with emotion annotation have been proposed, an utterance is annotated only with a speaker's emotion \citep{li-etal-2017-dailydialog, hsu-etal-2018-emotionlines} or a dialogue as a whole is annotated \cite{rashkin-etal-2019-towards}, all of which are not appropriate for enabling the above ability.

In this paper, we propose a method to build an emotion-annotated multi-turn dialogue corpus, which is necessary for developing a dialogue system that can recognize the emotion of an interlocutor and generate a response with an appropriate emotion.
We annotate each utterance in a dialogue with an \textbf{expressed emotion}, which a speaker put into the utterance, and an \textbf{experienced emotion}, which a listener felt when listening to the utterance.

\begin{figure}[t]
    \centering
    \includegraphics[width=\linewidth]{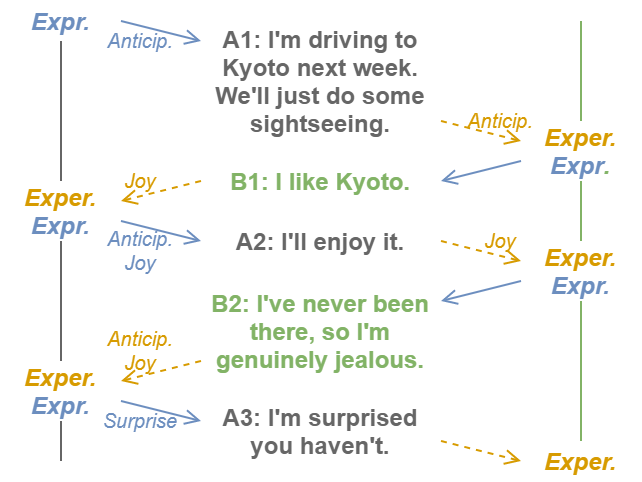}
    \caption{An example dialogue with expressed and experienced emotions.}
    \label{fig:ex}
\end{figure}

To construct a multi-turn dialogue corpus annotated with these emotions, we collect dialogues from Twitter and crowdsource their emotion annotation.
As a dialogue corpus, we extract tweet sequences where two people speak alternately.
For the emotion annotation, we adopt Plutchik's wheel of emotions \citep{plutchik1980general} as emotion labels and ask crowdworkers whether an utterance indicates each emotion label for expressed and experienced emotion categories.
Each utterance is allowed to have multiple emotion labels and has an intensity, strong and weak, according to the number of crowdworkers' votes.
We build a Japanese dialogue corpus as a testbed in this paper, but our proposed method can be applied to any language.

Using the above method, we constructed a Japanese emotion-tagged dialogue corpus consisting of 3,828 dialogues and 13,806 utterances.\footnote{We will release our corpus and code at \url{https://github.com/nlp-waseda/expr-exper-emo}.}
Statistical analysis of the constructed corpus revealed the characteristics of words for each emotion and the relationship between expressed and experienced emotions.
We further conducted experiments to recognize expressed and experienced emotions using BERT \citep{devlin-etal-2019-bert}.
We defined the task of emotion recognition as regression and evaluated BERT's performance using correlation coefficients.
The experimental results showed that it was more difficult to infer experienced emotions than expressed emotions, and that
multi-task learning of both emotion categories improved the overall performance of emotion recognition.
From these results, we can see that expressed and experienced emotions are different, and that it is meaningful to annotate both.
We expect that the constructed corpus will facilitate the study on emotion recognition in dialogue and emotion-aware response generation.

\section{Related Work}
\label{sec:work}

\subsection{Emotion-Tagged Corpora}
\label{ssec:work:emo}

Many non-dialogue corpora annotated with emotions have been constructed.
EmoBank \citep{buechel-hahn-2017-emobank} is a corpus of social media or reviews with emotion annotation.
They annotate sentences with the emotions of a person who read them and a person who wrote them.
% Also, they adopt not a categorical model but a dimensional model of emotions called VAD.
WRIME \citep{kajiwara-etal-2021-wrime} is an emotion-annotated corpus in Japanese, where SNS posts are tagged with both \textit{subjective} and \textit{objective} emotions.
The concept of this corpus is similar to EmoBank.
However, they emphasize the subjectivity of annotation and ask writers to annotate their own sentences with emotions.
Furthermore, EmoInt \citep{mohammad-bravo-marquez-2017-emotion} aims at the task of detecting emotion intensity.
They annotate Twitter posts with anger, fear, joy, and sadness and give each emotion a real value between 0 and 1 as the intensity level.

Some corpora are tagged with non-emotional factors, along with emotions.
EmotionStimulus \citep{10.1007/978-3-319-18117-2_12} and GroundedEmotions \citep{8273642} are corpora that focus on the reason for an expressed emotion.
The former uses FrameNet to detect a cause, while the latter treats weather and news as external emotion factors.
In terms of emotion labels, the two corpora adopt seven emotions (Ekman's six emotions \citep{doi:10.1080/02699939208411068} and shame) and two emotions (only happiness and sadness), respectively.
In StoryCommonsense \citep{rashkin-etal-2018-modeling}, a series of sentences comprising of a short story is tagged with \textit{motivation} and \textit{emotional reaction} for each character.
For emotion labels, they use some theories of psychology, including Plutchik's wheel of emotions \citep{plutchik1980general}.

None of the above corpora, however, are relevant to dialogue.
StoryCommonsense is similar to ours but differs in that characters in a story are annotated instead of speakers' utterances.

\subsection{Dialogue Corpora}
\label{ssec:work:dial}

Several dialogue corpora annotated with emotions are available.
DailyDialog \citep{li-etal-2017-dailydialog} is one collected from educational websites and tagged with emotions and intentions.
EmotionLines \citep{hsu-etal-2018-emotionlines} is a multi-turn dialogue corpus with annotation of emotions.
Both of them use seven labels for tagging: Ekman's six emotions \citep{doi:10.1080/02699939208411068} and an other/neutral emotion.
MELD \citep{poria-etal-2019-meld} is an extension of EmotionLines, tagged with not only emotions but also visual and audio modalities.
EmpatheticDialogues \citep{rashkin-etal-2019-towards} is a dialogue-level emotion-tagged corpus, considering two participants as a \textit{speaker} and a \textit{listener}, and tagged with the speaker's emotion and its context.

In EmpatheticDialogues, not each utterance but each dialogue is annotated, which is not suitable for recognizing emotional transition throughout a dialogue.
For Japanese, there is a Japanese version of EmpatheticDialogues called JEmpatheticDialogues \citep{sugiyama2021empirical}, which suffers from the same problem.
In this work, we conduct utterance-level annotation like DailyDialog and EmotionLines.
Although these corpora contain only the speaker's emotion (expressed emotion), we also annotate an utterance with the emotion of a person who hears it (experienced emotion).
Furthermore, while an utterance has only one emotion label in these corpora, we allow multiple emotion labels to be tagged per utterance and also consider their strength.

There are also some studies toward developing emotion-aware dialogue systems.
\citet{smith-etal-2020-put} propose three skills for a human-like dialogue system: recognizing emotions, using knowledge \citep{dinan2018wizard}, and considering personality \citep{zhang-etal-2018-personalizing}.
Furthermore, \citet{roller-etal-2021-recipes} build a dialogue system capable of blending these three skills.

\section{Corpus Building}
\label{sec:build}

\begin{table}[t]
    \centering
    \begin{tabular}{l|rr}
        \hline
        \textbf{Length} & \textbf{\# Dialogues} & \textbf{\# Utterances} \\
        \hline
        2 & 1,330 & 2,660 \\
        3 & 1,071 & 3,213 \\
        4 & 509 & 2,036 \\
        5 & 310 & 1,550 \\
        6 & 225 & 1,350 \\
        7 & 158 & 1,106 \\
        8 & 134 & 1,072 \\
        9 & 91 & 819 \\
        \hline
        2-9 & 3,828 & 13,806 \\
        \hline
    \end{tabular}
    \caption{The statistics of dialogues and utterances.}
    \label{tab:stats_len}
\end{table}

\begin{table}[t]
    \centering
    \begin{tabular}{@{}l|r@{ }r|r@{ }r@{}}
        \hline
        \multirow{2}{*}{\textbf{Label}} & \multicolumn{2}{c|}{\textbf{Expressed}} & \multicolumn{2}{c}{\textbf{Experienced}} \\
         & \textbf{Strong} & \textbf{Weak} & \textbf{Strong} & \textbf{Weak} \\
        \hline
        Anger & 430 & 1,349 & 124 & 870 \\
        Anticipation & 1,906 & 4,229 & 1,215 & 4,068 \\
        Joy & 1,629 & 3,672 & 1.553 & 4,549 \\
        Trust & 247 & 1,732 & 520 & 3,455 \\
        Fear & 252 & 942 & 123 & 846 \\
        Surprise & 602 & 2,018 & 434 & 2,798 \\
        Sadness & 1,227 & 2,936 & 889 & 3,037 \\
        Disgust & 476 & 1,979 & 186 & 1,535 \\
        \hline
        Any & 6,371 & 12,215 & 4,705 & 12,515 \\
        \hline
    \end{tabular}
    \caption{The statistics of utterances for each emotion label.}
    \label{tab:stats_emo}
\end{table}

\begin{figure}[t]
    \centering
    \includegraphics[width=\linewidth]{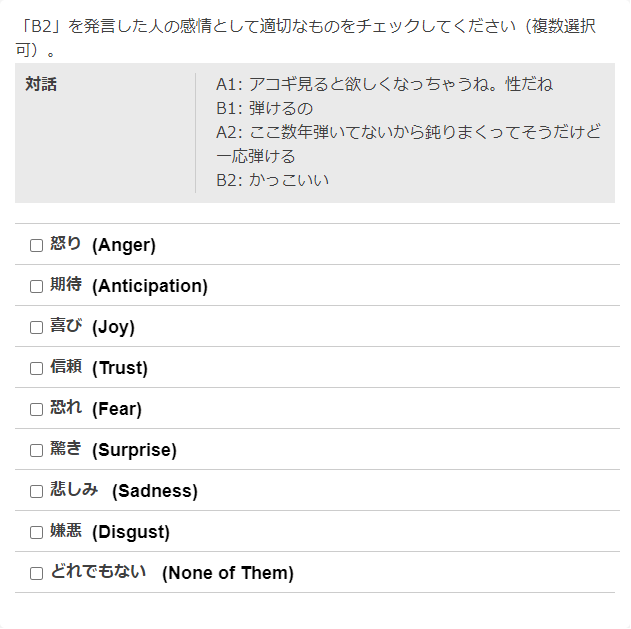}
    \caption{An example of the crowdsourced task. Checkboxes allow crowdworkers to select multiple emotions for an utterance.}
    \label{fig:yahoo}
\end{figure}

\begin{table*}[t]
    \centering
    \small
    \begin{tabular}{p{0.5\linewidth}|ll}
        \hline
        \textbf{Utterance} & \textbf{Expressed} & \textbf{Experienced} \\
        \hline
        A1: \begin{CJK}{UTF8}{ipxm}来週、車で京都行く 普通に観光してきます\end{CJK} (I'm driving to Kyoto next week. We'll just do some sightseeing.) & \{\textbf{Anticipation}, Joy\} & \{\textbf{Anticipation}\} \\
        B1: \begin{CJK}{UTF8}{ipxm}いいなぁ、京都\end{CJK} (I like Kyoto.) & \{Anticipation\} & \{Anticipation, \textbf{Joy}\} \\
        A2: \begin{CJK}{UTF8}{ipxm}楽しんできます\end{CJK} (I'll enjoy it.) & \{\textbf{Anticipation}, \textbf{Joy}\} & \{Anticipation, \textbf{Joy}\} \\
        B2: \begin{CJK}{UTF8}{ipxm}行ったことないから純粋に羨ましい\end{CJK} (I've never been there, so I'm genuinely jealous.) & \{Anticipation\} & \{\textbf{Anticipation}, \textbf{Joy}\} \\
        A3: \begin{CJK}{UTF8}{ipxm}ないんや意外\end{CJK} (I'm surprised you haven't.) & \{\textbf{Surprise}\} & \{Joy, Surprise\} \\
        \hline
    \end{tabular}
    \caption{An example dialogue annotated with expressed and experienced emotions by crowdsourcing. The labels in bold indicate strong emotions.}
    \label{tab:example}
\end{table*}

\begin{table*}[t]
    \centering
    \small
    \begin{tabular}{l|p{0.4\linewidth}p{0.4\linewidth}}
        \hline
        \textbf{Label} & \textbf{Expressed} & \textbf{Experienced} \\
        \hline
        Anger & \begin{CJK}{UTF8}{ipxm}糞, せる, マジだ\end{CJK} (shit, force, serious) & \begin{CJK}{UTF8}{ipxm}糞, うるさい, 居る\end{CJK} (shit, noisy, exist) \\
        Anticipation & \begin{CJK}{UTF8}{ipxm}教える, 願う, 待つ\end{CJK} (teach, hope, wait) & \begin{CJK}{UTF8}{ipxm}待つ, 楽しみだ, 強い\end{CJK} (wait, looking forward to, strong) \\
        Joy & \begin{CJK}{UTF8}{ipxm}楽しい, 嬉しい, おもろい\end{CJK} (joyful, glad, funny) & \begin{CJK}{UTF8}{ipxm}楽しい, 嬉しい, おもろい\end{CJK} (joyful, glad, funny) \\
        Trust & \begin{CJK}{UTF8}{ipxm}全然, 大丈夫だ, ちゃんと\end{CJK} (at all, all right, properly) & \begin{CJK}{UTF8}{ipxm}やすみ, 教える, 大事だ\end{CJK} (rest, teach, important) \\
        Fear & \begin{CJK}{UTF8}{ipxm}怖い, やばい, どう\end{CJK} (afraid, serious, how) & \begin{CJK}{UTF8}{ipxm}怖い, やばい, 危険だ\end{CJK} (afraid, serious, dangerous) \\
        Surprise & \begin{CJK}{UTF8}{ipxm}やばい, なんで, ？\end{CJK} (serious, why, ?) & \begin{CJK}{UTF8}{ipxm}居る, ビックリ, 年\end{CJK} (exist, surprise, year) \\
        Sadness & \begin{CJK}{UTF8}{ipxm}泣く, 痛い, 悲しい\end{CJK} (cry, hurt, sad) & \begin{CJK}{UTF8}{ipxm}泣く, 辛い, 痛い\end{CJK} (cry, hard, hurt) \\
        Disgust & \begin{CJK}{UTF8}{ipxm}悪い, 嫌いだ, 嫌だ\end{CJK} (bad, hate, dislike) & \begin{CJK}{UTF8}{ipxm}悪い, 気持ち, 嫌だ\end{CJK} (bad, surprise, dislike) \\
        \hline
    \end{tabular}
    \caption{Top-3 frequent words for each emotion label. An IDF filtering is applied to exclude common words.}
    \label{tab:stats_word}
\end{table*}

\begin{figure*}[t]
    \centering
    \begin{subfigure}[b]{0.45\linewidth}
        \centering
        \includegraphics[width=\linewidth]{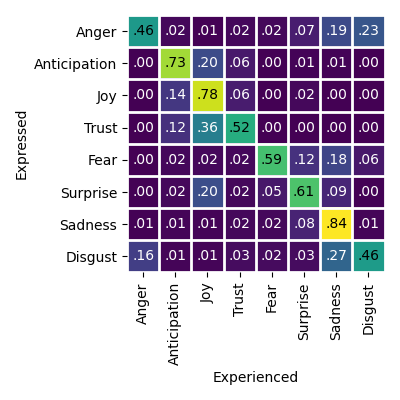}
        \caption{Expressed and experienced emotions (for a certain utterance).}
        \label{fig:mat_a}
    \end{subfigure}
    \hfill
    \begin{subfigure}[b]{0.45\linewidth}
        \centering
        \includegraphics[width=\linewidth]{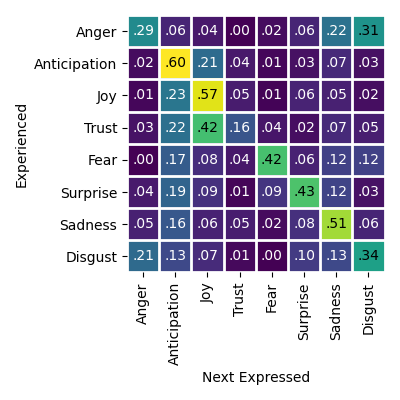}
        \caption{Experienced and next expressed emotions (for a certain person).}
        \label{fig:mat_b}
    \end{subfigure}
    \caption{The confusion matrices of the relationship between expressed and experienced emotions. In this analysis, we focus on only the strong labels. Note that the matrices' elements are normalized in the row direction.}
    \label{fig:mat}
\end{figure*}

\begin{figure*}[t]
    \centering
    \begin{subfigure}[b]{0.45\linewidth}
        \centering
        \includegraphics[width=\linewidth]{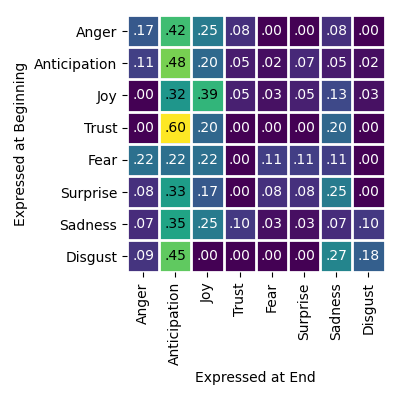}
        \caption{Expressed emotions at the beginning and end of dialogue.}
        \label{fig:mat_end_a}
    \end{subfigure}
    \hfill
    \begin{subfigure}[b]{0.45\linewidth}
        \centering
        \includegraphics[width=\linewidth]{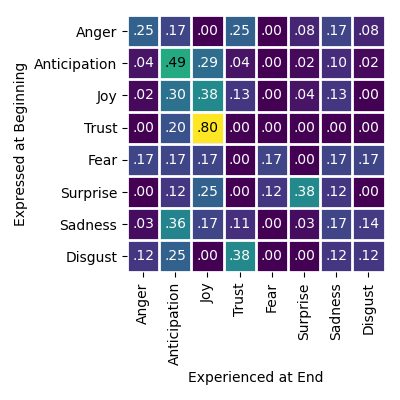}
        \caption{Expressed emotions at the beginning and experienced emotions at the end of dialogue.}
        \label{fig:mat_end_b}
    \end{subfigure}
    \caption{The confusion matrices for the emotion labels at the beginning and end of dialogue. In this analysis, we consider only the emotions of a person who begins a dialogue. Note that the targets are limited to the dialogues containing six to nine utterances, and the elements are normalized in the row direction.}
    \label{fig:mat_end}
    % \vspace*{-1.5ex}
\end{figure*}

\subsection{Dialogue Collection}
\label{ssec:build:dial}

We collect dialogue texts from Twitter by considering the interaction between tweets and their replies by two users as a dialogue.
To improve the text quality, we exclude tweets that contain images or hashtags and set the maximum number of utterances included in a dialogue to nine.
We also apply several filters: excluding dialogues that contain special symbols, emojis, repeated characters, and utterances that are too short.
Note that the reason why we exclude emojis is that they are relatively explicit emotional factors, and we intend to analyze emotions implied from usual textual expressions.

We collected Japanese dialogues using this method.
The numbers of dialogues and utterances are shown in Table~\ref{tab:stats_len}.
We obtained 3,828 dialogues that correspond to 13,806 utterances in total.
Regarding the length of dialogues, the number of dialogues tends to decrease as that of utterances per dialogue increases.

\subsection{Emotion Annotation}
\label{ssec:build:emo}

We adopt Plutchik's wheel of emotions \citep{plutchik1980general} as annotation labels.\footnote{Ekman's six emotions \citep{doi:10.1080/02699939208411068} and Plutchik's wheel of emotions \citep{plutchik1980general} are commonly used in emotion-tagged corpora. Preliminary experiments by crowdsourcing showed that the latter is more appropriate for our crawled dialogues. In this work, therefore, we use eight emotions by \citet{plutchik1980general}.}
Specifically, our annotation labels consist of eight emotions: anger, anticipation, joy, trust, fear, surprise, sadness, and disgust.
We annotate each utterance with two emotion categories: an expressed emotion, which is expressed by a speaker of the utterance, and an experienced emotion, which is experienced by a listener of the utterance.
In other words, an utterance is annotated with both subjective and objective emotions, which is similar to EmoBank \citep{buechel-hahn-2017-emobank} for non-dialogue texts.
By annotating expressed and experienced emotions, we can trace the changes in the emotion surrounding both an utterance and a participant in a dialogue.

As a crowdsourcing platform, we use Yahoo! Crowdsourcing.\footnote{\url{https://crowdsourcing.yahoo.co.jp/}}
By showing the target utterance and its context, we ask seven workers whether the target utterance has a specified emotion or not about each emotion label for expressed and experienced emotion categories.
For the expressed emotions, we ask which emotion a speaker expressed when saying the utterance.
For the experienced emotions, we ask which emotion a listener experienced when hearing the utterance.
Workers are allowed to select multiple emotion labels or none of them.
An interface of the crowdsourcing task for expressed emotions is shown in Figure~\ref{fig:yahoo}.

Because a view of expressed and experienced emotions can vary among annotators, we employ many workers per an utterance and aggregate their votes to obtain highly reliable annotations.\footnote{For the expressed emotion, we ask workers to annotate the emotion of the speaker of an utterance.
This annotation, however, is not strictly what a speaker had in mind but what the workers think a speaker would want to express, which can be considered \textit{objective} in some sense.
Having truly subjective annotation as the expressed emotion, like \citet{kajiwara-etal-2021-wrime}, is our future work.}
We consider strength for each emotion according to the number of workers' votes; emotions selected by more than half of the workers are regarded as strong, and ones selected by more than a quarter are regarded as weak.
Note that the set of strong emotions is a subset of the set of weak ones.
We expect that providing the emotions with intensity enables us to handle their granularity.

We applied the above emotion annotation method to our dialogue corpus.
The number of utterances for each emotion is shown in Table~\ref{tab:stats_emo}.
For the expressed emotion, 46.15\% and 88.48\% of the utterances are tagged with at least one strong and weak emotion, respectively.
For the experienced emotion, the percentages are 34.08\% and 90.65\%, respectively.
Approximately 90\% of the utterances are accompanied by one or more emotion labels, and thus our corpus is consequently suitable for recognizing emotions in dialogues and analyzing their changes.
In contrast to ours, for example, less than 20\% of utterances are tagged with a specific emotion in DailyDialog \citep{li-etal-2017-dailydialog}.
Hence it is difficult to analyze emotion changes using such corpora with a small amount of emotion annotation.
In addition, we can see a bias among the emotion labels for both expressed and experienced emotions, with more instances of anticipation and joy and fewer instances of trust and fear.
An example of a dialogue with the annotation is shown in Table~\ref{tab:example}.

\section{Corpus Analysis}
\label{sec:analysis}

\subsection{Frequent Words for Emotion Categories and Labels}
\label{ssec:analysis:word}

To investigate the characteristics of utterances with different emotions, we count words for each strong emotion label in our corpus.
In this analysis, we identify words by the Japanese morphological analyzer Juman++ \citep{tolmachev-etal-2018-juman}.
To exclude common words likely to appear for all emotions, we apply an IDF filtering.
Specifically, words with IDF less than half of the maximum are ignored.

Top-3 words appearing for strong emotion labels are shown in Table~\ref{tab:stats_word}.
The same words tend to appear in the two emotion categories for joy and sadness.
In contrast, the frequent words in the two categories are different for anticipation, trust, and surprise.

\subsection{Relationship Between Expressed and Experienced Emotions}
\label{ssec:analysis:mat}

We annotated utterances with the expressed and experienced emotions.
Here, we focus on the relationship between these two emotion categories.
Specifically, we investigate the following two relationships:
\begin{enumerate}[a.]
    \item The expressed emotion and the experienced emotion for the same utterance (different persons).
    \item The experienced emotion for an utterance and the expressed emotion for the next utterance (the same person).
\end{enumerate}

The confusion matrices for the strong emotion labels are shown in Figure~\ref{fig:mat}, where the elements are normalized in the row direction.
First, diagonal components of the two confusion matrices have large values, indicating that the same emotions are likely to occur both for the same utterance and for the same person.
Figure~\ref{fig:mat_a} shows that people are likely to experience joy for an utterance of anticipation, trust, and surprise in addition to the same emotion.
People also tend to experience disgust and sadness for anger and disgust, respectively.
Figure~\ref{fig:mat_b} shows that after experiencing trust, people are more likely to express joy than trust.
For an anger experience, people are more likely to express disgust than anger.
Figures~\ref{fig:mat_a} and \ref{fig:mat_b} reveal that the relationship of sadness is particularly different.
For a certain utterance, sadness makes the other person feel sad in most cases, but for a certain person, anticipation in addition to sadness can be expressed after experiencing sadness.
We speculate that when a person experiences sadness from the interlocutor, the person brings an utterance with anticipation to comfort them.

\subsection{Emotions at the Beginning and End of a Dialogue}
\label{ssec:analysis:bar}

To analyze the emotion changes through a dialogue, we compare emotions at the beginning and end of a dialogue.
In other words, we see how the emotions of a person who starts the dialogue change through the dialogue.
In this analysis, we focus on the following two relationships:
\begin{enumerate}[a.]
    \item The emotion expressed first and the emotion \textit{expressed} last by the same person.
    \item The emotion expressed first and the emotion \textit{experienced} last by the same person.
\end{enumerate}

The confusion matrices for the strong emotion labels are shown in Figure \ref{fig:mat_end}.
The targets are limited to dialogues containing six to nine utterances to analyze the emotion changes in long dialogues.
Figure~\ref{fig:mat_end_a} shows that a speaker of the first utterance is likely to finally express anticipation and joy regardless of the first emotion.
A speaker who first expresses surprise can express sadness through the dialogue. 
Figure~\ref{fig:mat_end_b} also shows that the first speaker can experience anticipation at the end of a dialogue.
A person who first expresses anger and disgust tends to finally experience trust.
From these two figures, we can see that a dialogue causes a person who first expresses fear to finally feel either a positive or negative emotion.

\section{Experiments}
\label{sec:exp}

\subsection{Model Setup}
\label{ssec:exp:model}

We conduct experiments on expressed and experienced emotion recognition using our corpus.
We solve a regression task of each emotion intensity for an utterance with its context for the emotion recognition task.
We assign 0, 1, and 2 for none, weak, and strong emotion labels, respectively, and let a model regress these values for each emotion.
As such, we train two separate models for expressed and experienced emotions with the mean squared error loss:
\begin{align}
    \mathcal{L} = \frac{1}{N K} \sum_{i=1}^N \sum_{j=1}^K (y_{ij} - t_{ij})^2,
\end{align}
where $N$ is the number of samples and $K$ is the number of emotion labels.
$y_{ij}$ is the output from the model for the $j$th label of the $i$th sample, and $t_{ij}$ is its gold label.

We adopt a Japanese pre-trained BERT model and fine-tune it.
We compare two pre-trained models from Kyoto University\footnote{\url{https://nlp.ist.i.kyoto-u.ac.jp/?ku_bert_japanese}} and one from NICT\footnote{\url{https://alaginrc.nict.go.jp/nict-bert/index.html}}.
We use the WWM and BPE versions for Kyoto University's and NICT's BERT models, respectively.

Input utterances are segmented into words with Juman++ \citep{tolmachev-etal-2018-juman} and tokenized into subwords by applying BPE.
We join utterances with \texttt{[SEP]} and append \texttt{[CLS]} and \texttt{[SEP]} to the beginning and end, respectively.
As there are two participants in a dialogue, we give each utterance a segment ID of 0 or 1.
It provides the models with the information about the speaker of an utterance.
Based on a series of utterances joined with \texttt{[SEP]}, we predict an emotion label for the last utterance.
The vector corresponding to \texttt{[CLS]} is passed to a fully-connected layer, and an eight-dimensional vector representing the eight emotions is obtained.
Each of the elements is supposed to regress the intensity of each emotion.

Since we are dealing with a regression task, Pearson's and Spearman's correlation coefficients are used as evaluation metrics.
The dialogues in our corpus are split into 8:1:1, corresponding to training, validation, and test sets.
We fine-tune our models for three epochs and evaluate them on the test set.
The implementation of the models is based on HuggingFace Transformers\footnote{\url{https://huggingface.co/transformers/}}.
The models are trained using NVIDIA Tesla V100 SXM2 GPU.

\subsection{Results}
\label{ssec:exp:result}

\begin{table}[t]
    \centering
    \begin{tabular}{l|rr}
        \hline
        \textbf{Model} & \textbf{Expressed} & \textbf{Experienced} \\
        \hline
        Kyoto (base) & 58.84/44.33 & 53.60/41.84 \\
        Kyoto (large) & 60.85/45.16 & 55.09/42.94 \\
        NICT & \textbf{61.50}/\textbf{46.05} & \textbf{56.23}/\textbf{43.88} \\
        \hline
    \end{tabular}
    \caption{The results of regression for expressed and experienced emotions. The metrics are Pearson's and Spearman's correlation coefficients.}
    \label{tab:res_model}
\end{table}

\begin{table}[t]
    \centering
    \begin{tabular}{l|rr}
        \hline
        \textbf{Label} & \textbf{Expressed} & \textbf{Experienced} \\
        \hline
        Anger & 50.21/33.80 & 38.11/23.80 \\
        Anticipation & 62.76/\textbf{55.55} & 57.46/51.22 \\
        Joy & \textbf{67.25}/55.22 & \textbf{61.92}/\textbf{54.47} \\
        Trust & 41.15/36.69 & 43.91/40.48 \\
        Fear & 59.09/31.47 & 49.60/24.90 \\
        Surprise & 49.86/39.58 & 40.58/33.86 \\
        Sadness & 63.70/51.50 & 55.48/43.88 \\
        Disgust & 47.76/38.18 & 37.32/28.13 \\
        \hline
    \end{tabular}
    \caption{The correlation coefficients for each emotion label. The metrics are Pearson's and Spearman's correlation coefficients. The scores are from the NICT model that achieved the highest performance in Table~\ref{tab:res_model}.}
    \label{tab:res_emo}
\end{table}

\begin{table*}[t]
    \centering
    \small
    \begin{tabular}{p{.7\linewidth}|p{.1\linewidth}p{.1\linewidth}}
        \hline
        \textbf{Dialogue} & \textbf{Predicted} & \textbf{Gold} \\
        \hline
        \begin{minipage}[t]{\linewidth}\begin{CJK}{UTF8}{ipxm}A1: ゲームの検証してる人が検証してほしいことあれば言ってください的なこと言ってたから依頼したら無視されて悲しくなったのはいい思い出 (I have a good memory of a guy who was verifying a game and said if there was anything he wanted verified, please let him know, so I made a request and he ignored it, which made me sad.) \\ \colorbox[gray]{0.9}{B1:} それは悲しいね (That's sad.)\end{CJK}\end{minipage} & \textbf{Strong sadness} & \textbf{Strong sadness} \\
        \hline
        \begin{minipage}[t]{\linewidth}\begin{CJK}{UTF8}{ipxm}A1: youtubeでバーのマスターが氷砕いてる動画見てボーッとしてる (I've been watching videos of bar masters crushing ice on youtube and I'm in a daze.) \\ B1: なんかしてよ (Do something.) \\ A2: そのうちこういうときにツイキャスをしようかなと思っておる (One of these days I'm going to do a tweak for this.) \\ \colorbox[gray]{0.9}{B2:} 天才の発想 スマホでも見やすいから助かる (It's a genius idea, and it's easy to watch on my phone.)\end{CJK}\end{minipage} & Weak anticipation and \textbf{joy} & Strong \textbf{joy} and weak trust \\
        \hline
        \begin{minipage}[t]{\linewidth}\begin{CJK}{UTF8}{ipxm}A1: 今、部活終わって帰るとこやけど 雨やばいしかっぱ持ってきてないし 最悪 (I'm on my way home after club activities, but it's raining and I didn't bring my hat, so that sucks.) \\ \colorbox[gray]{0.9}{B1:} わたしも学校出た瞬間大雨降ってきた (I'm going back to school now, but it's raining really hard and I didn't bring my jacket.)\end{CJK}\end{minipage} & Strong surprise & Strong sadness \\
        \hline
    \end{tabular}
    \caption{Example dialogues with predicted and gold expressed emotions. The predicted emotion labels are taken from the predictions of the NICT model, which predicted an emotion label for the last utterance of each dialogue.}
    \label{tab:result_example}
\end{table*}

\begin{table}[t]
    \centering
    \begin{tabular}{l|rr}
        \hline
        \textbf{Train\textbackslash Test} & \textbf{Expressed} & \textbf{Experienced} \\
        \hline
        Expressed & 61.50/46.05 & 52.89/40.91 \\
        Experienced & 55.49/43.34 & 56.23/43.88 \\
        Multi-Task & \textbf{62.20}/\textbf{46.63} & \textbf{57.35}/\textbf{45.01} \\
        \hline
    \end{tabular}
    \caption{The results of multi-task learning with expressed and experienced emotions. The metrics are Pearson’s and Spearman’s correlation coefficients.}
    \label{tab:res_multi_u}
\end{table}

\begin{table}[t]
    \centering
    \begin{tabular}{@{}l|r@{\ \ }r@{}}
        \hline
        \textbf{Train\textbackslash Test} & \textbf{Experienced} & \textbf{Next Expressed} \\
        \hline
        Experienced & 54.62/43.47 & 29.53/25.46 \\
        Next Expressed & 43.32/35.27 & 33.91/28.31 \\
        Multi-Task & \textbf{55.75}/\textbf{49.50} & \textbf{35.17}/\textbf{30.49} \\
        \hline
    \end{tabular}
    \caption{The results of multi-task learning with experienced and next expressed emotions. The metrics are Pearson’s and Spearman’s correlation coefficients.}
    \label{tab:res_multi_s}
\end{table}

For the regression task defined in Section~\ref{ssec:exp:model}, the correlation coefficients for each model are shown in Table~\ref{tab:res_model}.
In terms of performance, the NICT model achieved the best score across all values.
For the values regarding expressed and experienced emotions, the performance of the experienced emotion is inferior to that of expressed emotion in all models.
This result indicated that it is more difficult to recognize the experienced emotion than the expressed emotion.

The correlation coefficients for each emotion inferred by the NICT model are shown in Table~\ref{tab:res_emo}.
For both the expressed and experienced emotions, the highest scores were achieved for \textit{anticipation} and \textit{joy}.
In contrast, the emotions with lower values were \textit{trust} and \textit{fear} for the expressed emotion and \textit{anger} and \textit{disgust} for the experienced emotion.
From Tables~\ref{tab:res_emo} and \ref{tab:stats_emo}, we can see that the larger the number of the samples for an emotion is, the higher the correlation coefficient becomes.
As a case study, we show example dialogues and their emotions predicted by the NICT model in Table~\ref{tab:result_example}.

\subsection{Multi-Task Learning}
\label{ssec:exp:multi}

Our analysis in Section~\ref{ssec:analysis:mat} indicated that there is a correlation between expressed and experienced emotions.
Therefore, we consider training a single model for recognizing both the emotion categories.
The information for solving the two similar tasks is expected to allow a model to improve the performance of each other \citep{liu-etal-2019-multi}.
We provide a model with two separate fully-connected layers for the tasks and train them simultaneously, where the inputs are the same as those in Section~\ref{ssec:exp:model}.
Here, the mean of the losses for expressed and experienced emotions is optimized:
\begin{align}
    \mathcal{L}_\text{multi-task} = \frac{\mathcal{L}_\text{expressed} + \mathcal{L}_\text{experienced}}{2}.
\end{align}
Based on Figures~\ref{fig:mat_a} and \ref{fig:mat_b}, we consider multi-task learning of expressed and experienced emotions for a certain utterance and a certain person.
For the relationship in a certain person, we use the experienced emotion of an utterance and the expressed emotion of the following utterance.
We also conduct experiments on the cases where the training and test sets are different from each other. In such a case, for example, expressed emotions are used for training, but experienced emotions are used for testing.

The correlation coefficients for an utterance and a speaker by multi-task learning are shown in Tables~\ref{tab:res_multi_u} and \ref{tab:res_multi_s}, respectively.
First, the scores when the training and test sets are different from each other are lower than those when they are the same.
This gap indicates the significance of annotating utterances with expressed and experienced emotions separately.
In all columns, the multi-task models achieved higher performance than the single-task models.
Especially, in Table~\ref{tab:res_multi_s}, the multi-task scores for both the two tasks are higher than the single-task baselines by one point.
In other words, expressed, experienced, and next expressed emotions have the information for helping the recognition of each other.

\section{Conclusion}

We proposed a method to build an emotion-tagged multi-turn dialogue corpus to help machines recognize emotional transition in a dialogue.
Dialogues between two speakers are collected from Twitter, and each utterance is annotated with emotions by crowdsourcing.
In the annotation process, we consider the emotions expressed by a speaker who said the utterance and the emotion experienced by a listener who heard the utterance. 
In addition, the labels are provided with their intensity, representing the granularity of emotions.

We built a Japanese emotion-tagged dialogue corpus and analyzed it.
The results showed the characteristics of words for each emotion, the correlation between the emotions about a certain utterance and speaker, and the tendency for speakers to become positive through a dialogue.
We also developed emotion recognition models for expressed and experienced emotions based on the Japanese pre-trained BERT models.
The experimental results indicated that it is more difficult to recognize a listener's emotion than a speaker's emotion.
Multi-task learning of expressed and experienced emotions improved the performance of the two emotion recognition tasks about an utterance and a speaker.

For our future work, we will tackle response generation based on predicted emotions.
With our corpus, a dialogue system is expected to predict which emotion it experiences from a given utterance and which emotion it should express for the next utterance.
Once such emotions are recognized, the dialogue system should be able to generate an appropriate response depending on the predicted expressed emotion.

The corpus in this work is annotated only with expressed and experienced emotions about an utterance.
In addition to the emotion annotation, we should also consider dialogue situations \cite{rashkin-etal-2019-towards}.
The cause of a dialogue or an utterance helps recognize a speaker's emotion and how it changes.
We can also consider non-emotional annotation, such as a dialogue's topic and an utterance's intention \cite{li-etal-2017-dailydialog}.
The relationship between emotions and non-emotional factors is also important for machines to better recognize a speaker's emotion.

\section*{Acknowledgements}

This work was supported by a joint research grant from LINE Corporation.

\section*{Ethical Considerations}

We built the dataset by collecting texts from Twitter and annotating them by crowdsourcing.
For crowdsourcing, we employed 3,847 workers.
It took approximately five minutes for a task of annotating 10 utterances.
Every worker was paid 4 JPY per 10 utterances, and in total, the built dataset costs 195,700 JPY.
Since the dataset was collected from Twitter, it may include contents that are harmful for some of the dataset or its application users. For building the dataset through the Twitter API and crowdsourcing, we did not include any sensitive information that allows personal identification.

The dataset or models trained on it enable downstream applications to infer the emotions of their users, resulting in facilitating communication between the users and the applications.
In terms of dialogue systems, this ability is considered valuable for both task-oriented and non-task-oriented dialogue systems.
For example, it assists the user in decision-making and solves the user's worry and trouble.
In contrast to such benefit, it is difficult for the model to infer the emotion accurately, with the relatively small dataset. Therefore, prediction errors by the model, especially for sensitive utterances or negative emotions, may bring harmful experiences on the users.

\bibliography{anthology,custom}
\bibliographystyle{acl_natbib}

% \appendix

% \section{Example Appendix}
% \label{sec:appendix}

% This is an appendix.

\end{document}